\documentclass[10pt,twocolumn,letterpaper]{article}

\usepackage{iccv}
\usepackage{times}
\usepackage{epsfig}
\usepackage{graphicx}
\usepackage{amsmath}
\usepackage{amssymb}
\usepackage{float}
\usepackage{soul}
\usepackage{pdfpages}

\newcommand{\smallpara}[1]{\noindent {\bf #1}}
\usepackage[pagebackref=true,breaklinks=true,letterpaper=true,colorlinks,bookmarks=false]{hyperref}

\iccvfinalcopy % *** Uncomment this line for the final submission

\def\iccvPaperID{} % *** Enter the ICCV Paper ID here

\ificcvfinal\fi
\begin{document}

%%%%%%%%% TITLE
\title{Surface Normals in the Wild}

\author{
	Weifeng Chen$^1$ \qquad   Donglai Xiang$^2$\thanks{Work done while a visiting student at the University of Michigan.} \qquad     Jia Deng$^1$ \\
$^1$University of Michigan, Ann Arbor, USA\\
$^2$Tsinghua University, Beijing, China\\
}

\maketitle

\begin{figure*}[t]
	\begin{center}
		\includegraphics[width=0.7\linewidth]{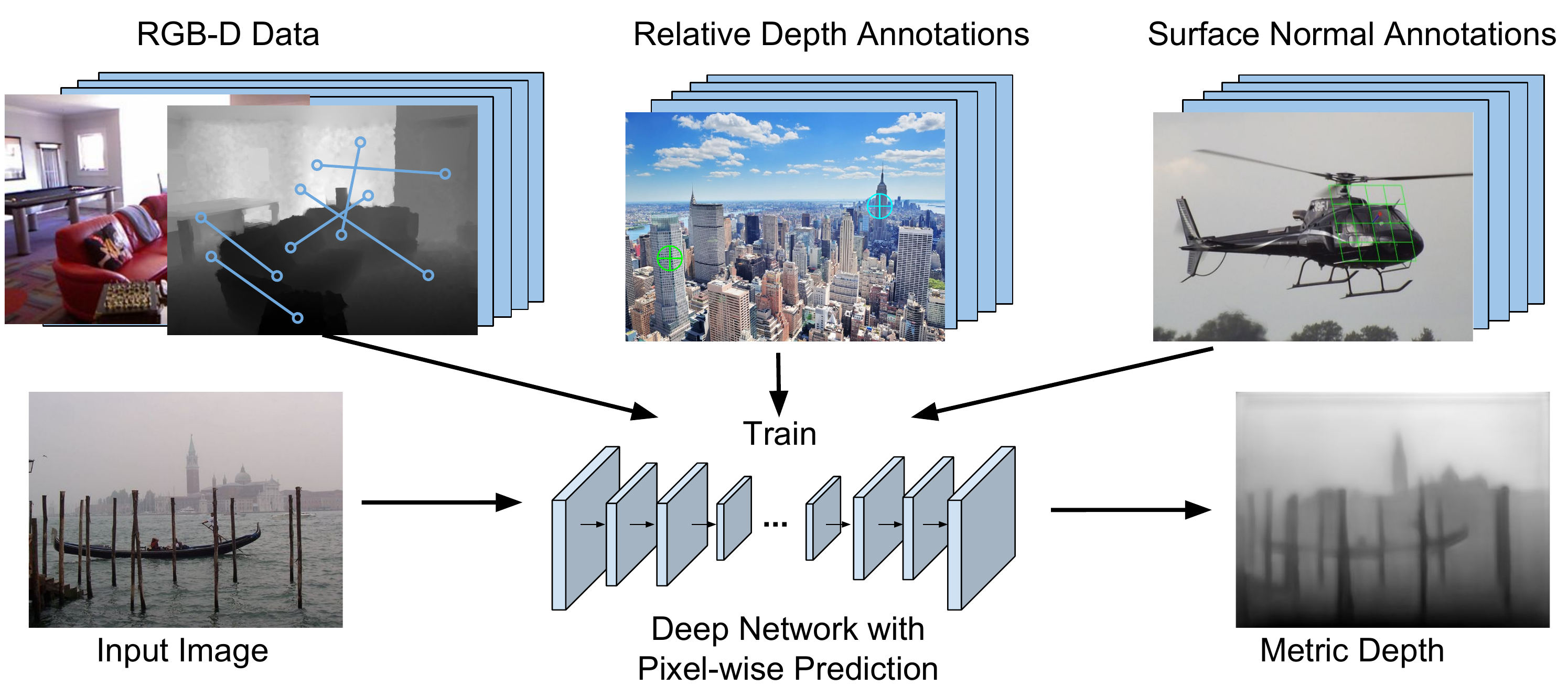}
	\end{center}
	\vspace{-3mm}
	\caption{Building on top of the work of Chen et al.~\cite{chen2016single}, we crowdsource annotations of surface normals and use the collected surface normals to help train a better depth prediction network.}
	\label{fig:teaser}
	\vspace{-3mm}
\end{figure*}

%%%%%%%%% ABSTRACT
\begin{abstract}
We study the problem of single-image depth estimation for images in the wild. 
We collect human annotated surface normals and use them to train a neural network that
directly predicts pixel-wise depth. We propose two novel loss functions for training with
surface normal annotations. Experiments on NYU Depth and our own dataset demonstrate that
our approach can significantly improve the quality of depth estimation in the wild. 

\end{abstract}

%%%%%%%%% BODY TEXT
\section{Introduction}

Single-image depth estimation is an important computer vision problem that has the potential
to majorly boost higher-level tasks such as object recognition and scene
understanding. However, despite extensive research~\cite{liu2015deep, eigen2014depth, li2015depth,
baig2015coupled, eigen2015predicting, ladicky2014pulling, wang2015towards,
wang2015designing, Karsch:TPAMI:14,chakrabarti2016depth,roymonocular,ranftldense,
bansal2016marr}, single-image depth estimation remains difficult. In particular, it
remains difficult to estimate depth for unconstrained images of arbitrary scenes, because, as prior work~\cite{chen2016single} has pointed out, existing RGB-D
datasets used to train current systems were collected by depth sensors. As a result, they
consist of a few specific types of indoor and outdoor scenes. Systems trained on these
datasets thus cannot generalize to images ``in the wild'' of arbitrary scenes and
compositions. 

Recent work by Chen et al.~\cite{chen2016single} made an attempt to
estimate depth for images ``in the wild'': they collected human annotations of relative
depth---the depth ordering of two points---for
random Internet images and use the annotations to train a deep network that directly predicts
metric depth. Chen et al. showed that it is possible to improve depth estimation for images
in the wild by using human annotations of depth. In particular, they showed that while it
is difficult to obtain absolute metric depth (per-pixel depth values) from humans, it is
nonetheless feasible to collect \emph{indirect, qualitative} depth annotations such as
relative depth, and use such annotations to learn to estimate metric depth. This strategy
does not rely on depth sensors and can work with arbitrary images; it thus has the
potential to significantly advance depth estimation in the wild. 

\begin{figure}[t]
\begin{center}

   \includegraphics[width=0.8\linewidth]{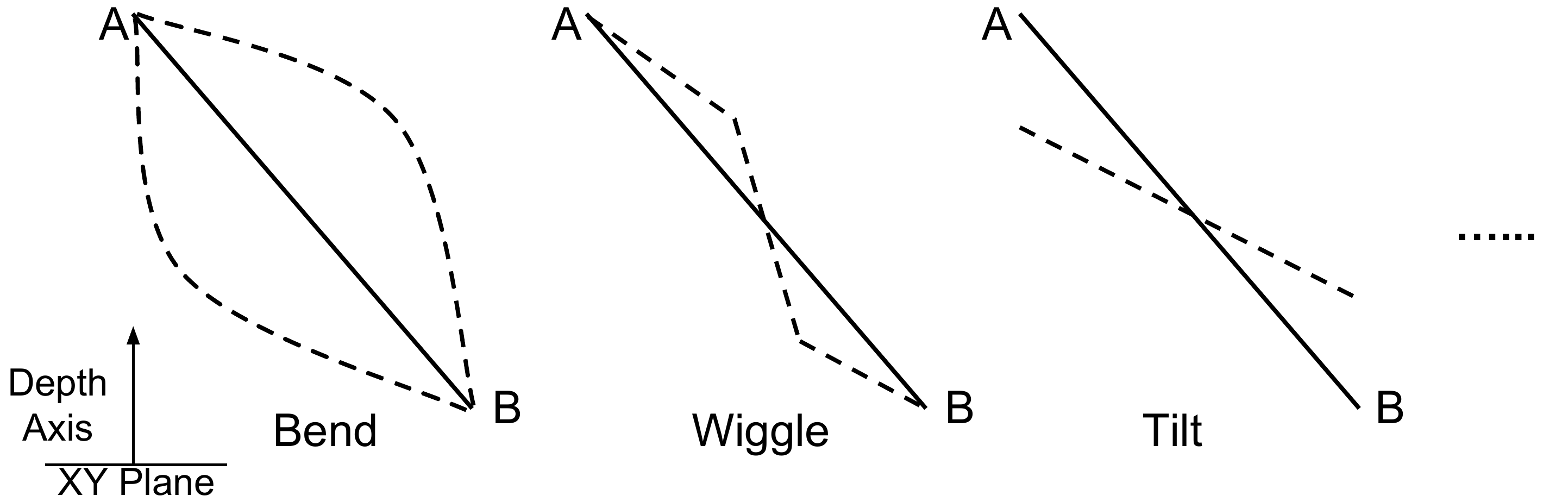}
\end{center}
\vspace{-3mm}
   \caption{Ambiguities of relative depth annotation. Bending, wiggling, or tilting a 3D surface from solid line configuration to dotted line configuration does not change the ordinal relation that point A is farther away from the camera than point B.}
\label{fig:ambiguities}
\end{figure}

One limitation of the work by Chen et al.~\cite{chen2016single}, however, is that annotations of relative depth
 do not capture all information that is perceptually important. 
 In particular, relative depth is invariant to monotonic transformations of
metric depth, meaning that there can be two scenes that are perceptually very different 
 yet are indistinguishable in terms of relative depth. For example, it
is possible to bend, wiggle, or tilt a straight line without affecting relative depth
(Fig.~\ref{fig:ambiguities}). In other words, relative depth does not capture important perceptual 
properties such as continuity, surface orientation, and curvature. As a result,  systems
trained on relative depth will not necessarily recover depth that is perceptually
faithful in all aspects. 

In this paper, we build on the work of Chen et al.~\cite{chen2016single} and address the limitation by introducing
an additional type of indirect, qualitative depth annotation---surface normals. Surface
carries important information on 3D geometry: they encode the local orientation of
surfaces and the derivatives of depth. In fact, given dense surface normals, it is
possible to recover full metric depth up to scaling and translation. This suggests that
annotations of surface normals can eliminate the ambiguities in relative depth and result in better depth
estimation. In addition, it has been well documented in human vision research that humans perceive surface orientation with
a remarkable degree of consistency~\cite{koenderink1992surface}. This suggests that it could
be feasible to collect human annotations for images in the wild. 

We consider
two questions: how to crowdsource annotations of surface normals, and how to use  surface
normal annotations to help train a network that predicts per-pixel metric depth. 
To crowdsource surface normals, we develop a UI that allows a user to annotate a surface normal
by adjusting a virtual arrow and a virtual tangent plane. This UI allows human annotators
to reliably estimate surface normals. With this UI we introduce a dataset called
``Surface Normals in the Wild'' (SNOW), which consists of surface normal annotations
collected from 60,061 Flickr images.

To incorporate surface normal
annotations into training, we develop two novel loss functions to train a deep network that directly
predicts metric depth. The first loss function is based on directly comparing normals,
that is, computing the angular difference between the ground truth normals
and the normals derived from the predicted depth. The second loss function is based on
comparing depth derivatives, i.e.\@,
computing the discrepancy between the derivative of the predicted depth and the
derivative given by the ground truth normals.  We show that each approach incurs
its own trade-offs and emphases on different aspects of depth quality, and should be chosen based on particular
applications.

Our main contributions are (1) a new dataset of crowdsourced surface normals for images in
the wild and (2) two distinct approaches of for using surface normal annotations 
 to train a deep network that directly predicts per-pixel
metric depth. Experiments on both NYU Depth~\cite{silberman2012indoor} and SNOW
demonstrate that surface
normal annotations can significantly improve the quality of depth estimation.

\section{Related work}

\noindent
\textbf{Datasets with depth and surface normals}
Prior works on estimating depth or surface normals have mostly used NYU Depth
\cite{silberman2012indoor} , Make3D~\cite{saxena20083}, 
KITTI~\cite{geiger2013vision}, or ScanNet~\cite{dai2017scannet}. Although these datasets provide highly accurate depth,
as pointed out by Chen et al.~\cite{chen2016single} they are limited to specific types of
scenes. The same limitation applies to synthetic datasets such as MPI Sintel~\cite{Butler:ECCV:2012} and the dataset by~\cite{richter2016playing} because the 3D content had to be manually created. 
The Depth in the Wild (DIW) dataset introduced
by Chen et al.~\cite{chen2016single} takes a major step toward including arbitrary scenes in the
wild.  However, DIW provides only relative depth annotations, which lack information on
many essential 3D properties such as surface normals. We build upon DIW and introduce 
a new dataset of crowdsourced surface normals for images in the wild. 

%discuss open surface. 
Open Surfaces~\cite{bell13opensurfaces} is a large dataset of images
with annotations of surface properties including surface normals and material. However, open
Surfaces is not suitable for depth estimation in the wild: it contains only images of indoor
scenes. In addition, it only has surface normals for planar surfaces, whereas
our dataset has no such restriction. 

\noindent
\textbf{Depth and surface normals from a single image} There has been a large body of
work on estimating depth and/or surface normals from a single image~\cite{liu2015deep, eigen2014depth, 
li2015depth, baig2015coupled, eigen2015predicting,laina2016deeper,
ladicky2014pulling, wang2015towards, wang2015designing, Karsch:TPAMI:14,
BarronTPAMI2015}. All these methods use dense ground truth depth or normals during
training, except the work of Zoran et al~\cite{zoran2015learning} which uses relative
depth for training. They all have difficulty generalizing to images in the wild due to the
limited scene diversity of the existing datasets that were acquired by depth sensors. 

Chen et al.~\cite{chen2016single} instead use crowdsourced relative depth for training,
using indirect depth human annotations to get around the limitations of depth sensors. Our
work goes beyond the work of Chen et al. by exploring surface normals. 

Two other recent works~\cite{garg2016unsupervised,xie2016deep3d} have also leveraged
indirect supervision of depth. In particular, they have used pairs of stereo images to impose constraints
on the predicted depth, e.g.\@ the depth estimated from the left image should be
consistent with the depth estimated from the right image as dictated by epipolar
geometry~\cite{garg2016unsupervised}. 

Chakrabarti et al.~\cite{chakrabarti2016depth} trained a network that simultaneously predicts distributions of
depth and distributions of depth derivatives at each pixel location. Then they used a
global optimization method to recover a
single depth map that is most consistent with the predictions.
Our work differs in two ways. First, the only output
of our network is a depth map. Our network does not
directly predict surface normals or depth derivatives, and thus there is no need for additional
optimization steps to harmonizing the outputs. Second, we do not use dense ground truth metric
depth in training. Our ground truth annotations are sparse and involve only relative depth and/or
surface normals.

\noindent
\textbf{Surface normals in 3D reconstruction}
Surface normals have played important roles in many 3D reconstruction systems. 
For example, surface normals have been used to infer 3D
models~\cite{kushal2013single}, create watertight 3D surfaces~\cite{Kazhdan:2006}, regularize planar object reconstruction~\cite{wang2016surge}, and to aid multi-view reconstruction~\cite{gallianijust} and structure from motion~\cite{ikehata2016panoramic},
or depth estimation~\cite{hane2015direction}. In our approach, surface normals are used in
training only; the network directly predicts depth, without explicitly producing surface
normals. 

\section{Dataset construction} 

Similar to the Depth in the Wild (DIW) dataset by Chen et al.~\cite{chen2016single}, we source our images 
from Flickr using random keywords from an English dictionary. For each image, we extract
the focal length of the camera from the EXIF metadata---the focal length is needed for
determining the amount of perspective distortion when we visualize a surface normal on top
of an image in our UI. 

\begin{figure}[t]
\begin{center}

   \includegraphics[width=0.8\linewidth]{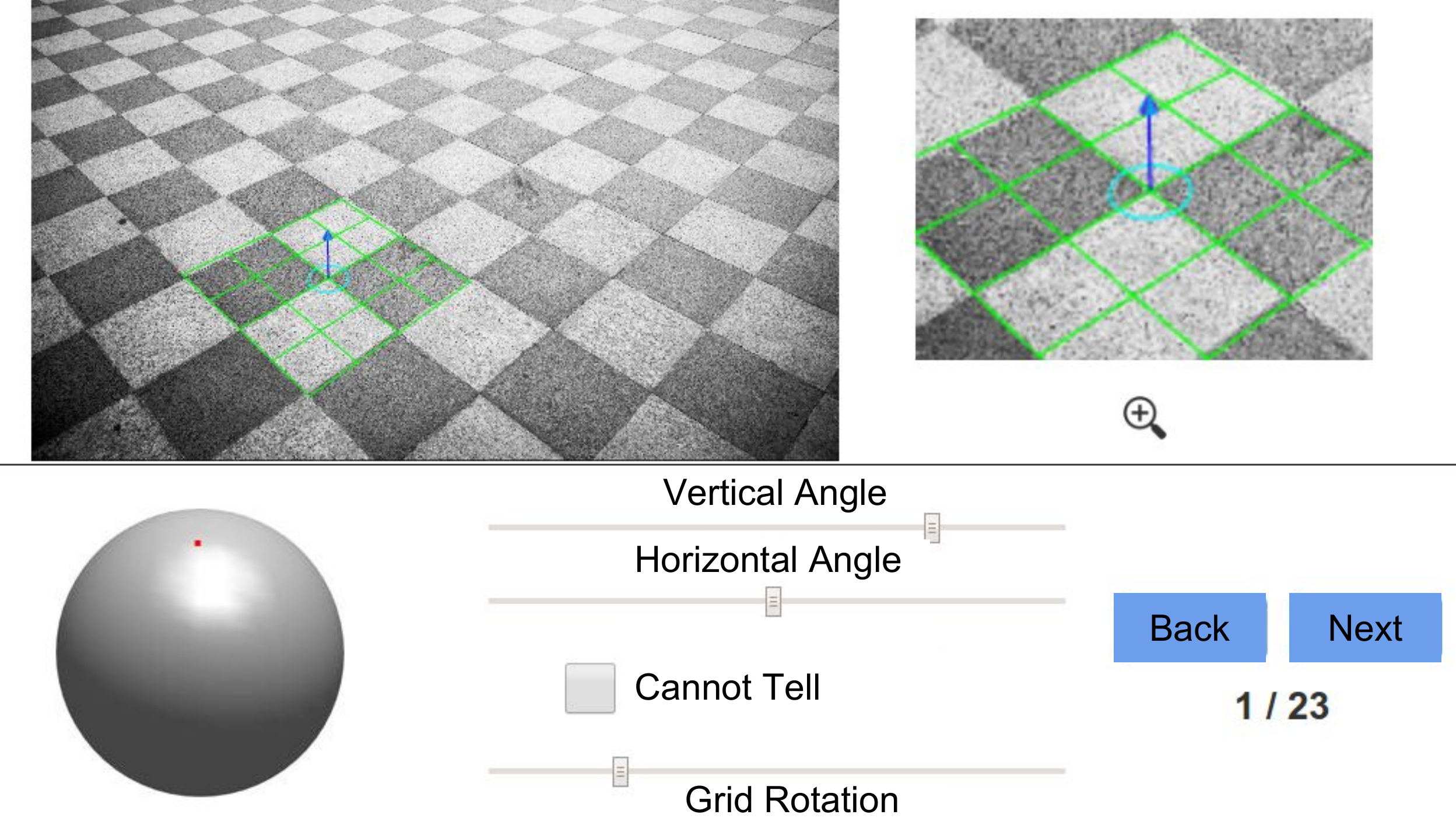}
\end{center}
   \caption{The annotation UI we use for data collection. The query image is displayed on the top left with the keypoint highlighted. A zoom-in view centered at the keypoint is displayed on the top right to help the worker see the details better. Workers then click on the sphere and adjust the slider bars to annotate the surface normal.}
\label{fig:amt_ui}
\end{figure}

To collect surface normal annotations, we present a crowd worker with 
an image and a highlighted location (Fig.~\ref{fig:amt_ui}). The worker then draws a
surface normal using a set of controls: she can pick a point on a sphere, or use two
slider bars to adjust the angles (there are two degrees of freedom). The surface normal is
visualized as an arrow originating from a 2D grid that represents the tangent plane. Both
the arrow and the 2D grid are rendered taking into account the focal length extracted from
the image metadata. This visualization is inspired by the gauge figures used in human vision
research~\cite{koenderink1992surface}; it helps the worker perceive the surface normal in 3D.

\begin{figure*}[t]
\begin{center}

   \includegraphics[width=\linewidth]{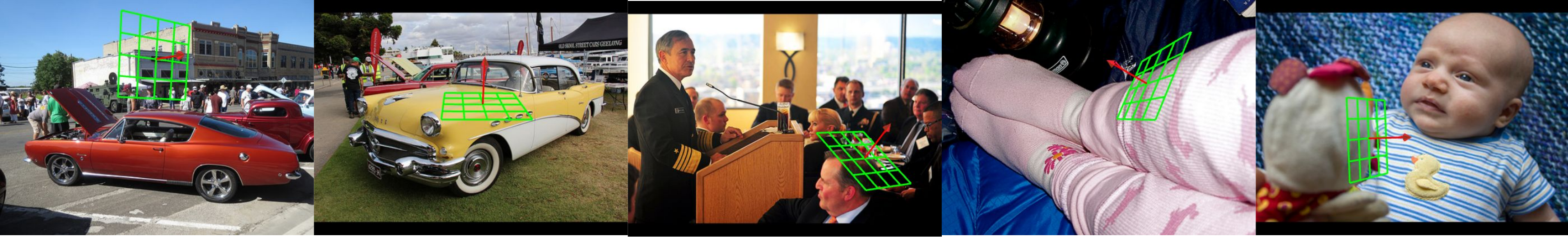}
\end{center}
\vspace{-3mm}
   \caption{Some examples of the final surface normal annotations we gather for the SNOW dataset. The green grid denotes the tangent plane, and the red arrow denotes the surface normal. For best visual effect, please view in color.}
\label{fig:example_annotation}
\vspace{-3mm}
\end{figure*}

For each image, we pick one random location uniformly from the 2D plane to have its
surface normal annotated. Following Chen et al.~\cite{chen2016single} we only pick one random location to minimize the correlation
between annotations. 

As the locations are randomly picked, some may fall onto areas where the surface normal is
hard to infer, especially when there is a large amount of clutter or texture, e.g.\@ tree
leaves in the distance or grass in a field. Surface normals may also be impossible to infer on 
regions such as the sky or a dark background (some examples are shown in the Appendix). In these cases a user can indicate that the surface normal
is hard to tell. 

We crowdsource the task through Amazon Mechanical Turk. We randomly inject
gold standard samples into the task to identify spammers. Each surface normal is annotated by two different workers. If
the two annotations are within 30 degree of each other, then we take the average of the
two (renormalized to a unit vector) as the final annotation; otherwise, we discard both annotations. 

Fig.~\ref{fig:example_annotation} shows some examples of the collected normals. In total,
we processed 210,000 images on Amazon Mechanical Turk and obtain  60,061 valid samples. On average, it takes about 15 seconds for a
 worker to annotate one surface normal. The average angular difference
between the two accepted annotation is 14.32$^{\circ}$.  This suggests that human annotations usually agree with each other quite well.

\subsection{Quality of human annotated surface normals}
An important question is how consistent and accurate the human annotations are. To study 
this, we collect human annotations of surface normals on a random sample of 113 NYU Depth~\cite{silberman2012indoor} images.
 Each surface normal is estimated
by three human annotators. We compare the human annotations with the
ground truth surface normals (derived from the Kinect ground truth depth). We measure the
Human-Human Disagreement (HHD) using the average angular difference between a human annotation
and the mean of multiple human annotations. We measure Human-Kinect Disagreement (HKD)
using the average angular difference between a human annotation and the Kinect ground truth.

We found that the Human-Human Disagreement on our sample is ($7.4^\circ$). This suggests
that human annotations are remarkably consistent between each
other. However, the Human-Kinect Disagreement is $32.8^{\circ}$ 
%(the \textbf{Overall} column in Tab.~\ref{tab:amt_exp_on_NYU}), 
which at first glance seems to suggest that human
annotations contain a large amount of systemic bias measured against the Kinect ground truth. However, a close inspection
reveals that most of the disagreement is a result of imperfect Kinect
ground truth rather than biased human estimation. 

One source of Kinect error is holes in the raw depth map. Some holes are due to specular or
reflective surfaces; others are due to the parallax caused by the RGB camera located
slightly away from the depth camera. The holes in the raw Kinect depth map are filled
through some heuristic post-processing. Such hole-filling is imperfect. It is especially
problematic at cluttered regions because it cannot recover the fine variations of depth
and as a result the derived normals will be inaccurate. 

Another source of Kinect error is imperfect normals computed from accurate depth. In this
experiment we used the official toolkit from the NYU Depth dataset~\cite{silberman2012indoor} to compute normals. Each normal is computed
by fitting a plane to a neighborhood of pixels. But this procedure tends to smooth out
 normals at or close to sharp normal discontinuities (e.g. at the intersection of two
planes or at occlusion boundaries). This problem is especially severe in cluttered regions where there are many such
discontinuities. But human estimation of normals is not susceptible to this issue. 

\begin{figure}[t]
\begin{center}
\includegraphics[width=\linewidth]{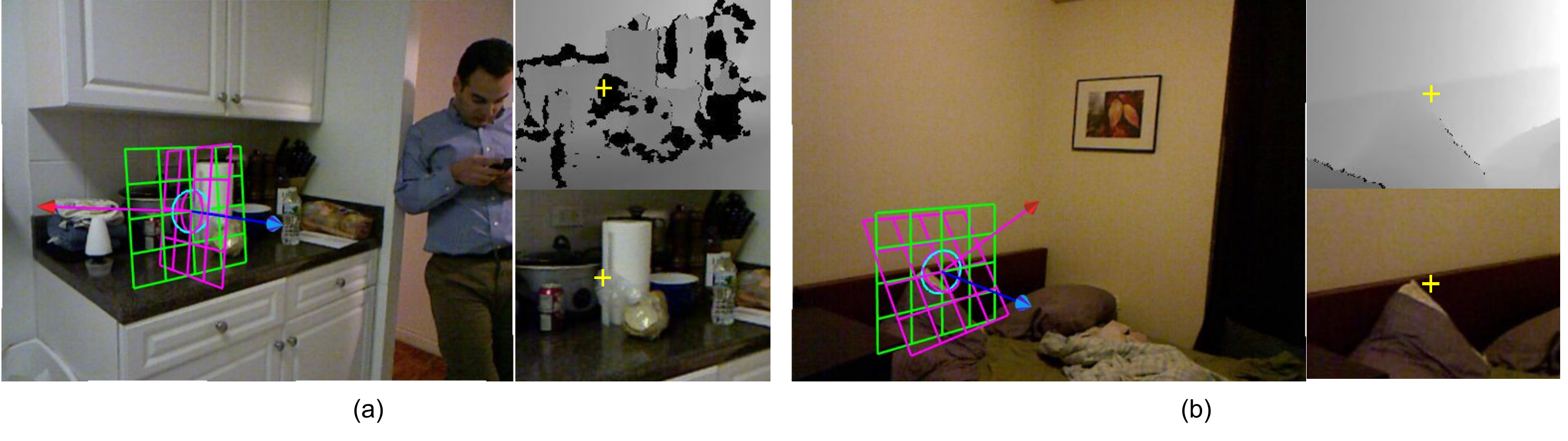}
\end{center}
\vspace{-3mm}
\caption{Examples of Kinect error. It shows annotations along with zoom-in views of depth map and 
RGB image around the keypoint (yellow cross). The red arrow with a purple mesh shows the Kinect ground-truth. 
Blue arrow and green mesh shows human annotations. (a) lies on a hole in the depth map which is 
caused by the transparent plastic bag. (b) lies near depth discontinuities. The surface normal in these
region cannot be reliably computed.  }
\label{fig:NYU_amt_failure_case}

\end{figure}

We manually inspected every image in our sample and found that 37\% of the cases can be
attributed to one of the two sources of Kinect error (holes or imperfect normal
calculation). Fig.~\ref{fig:NYU_amt_failure_case} shows examples of such cases. 
The Human-Kinect disagreement on these problematic cases is
44.32$^\circ$. Excluding these cases, the Human-Kinect disagreement is only
15.64$^\circ$. It is worth noting that in those cases of Human-Kinect disagreement,
humans remain remarkably consistent among themselves (average disagreement is
7.17$^\circ$). These results suggest that human annotations of surface normals
are of high quality. 

It is worth noting that due to the inherent ambiguity of
single-image depth estimation, we can never expect humans to match the accuracy of
depth sensors, which use more than a single image to recover depth. And in many
applications, especially those involving recognition, metric fidelity is not
essential. Consistency is the more important quality measure because it means that
there is a consistent representation (possibly biased) that we can hope to learn to estimate.

\section{Learning with surface normals}
\label{section:Learning_with_surface_normals}
Our goal is to train a deep neural network to perform depth prediction. We build our
method upon~\cite{chen2016single}, which uses relative depth as supervision during
training. The main idea from ~\cite{chen2016single} is to train a network using a
loss function that penalizes the inconsistency between the predicted depth and the ground
truth relative depth (ordinal relations between pairs of points). 
We propose to incorporate surface normals as additional supervision. This translates to a
loss function that encourages the predicted depth to be consistent with
both the ground truth relative depth and the ground truth surface normals. 

Formally, let $I$ be a
training image with $K$ relative depth annotations and $L$ surface normal
annotations. Using the same notations of ~\cite{chen2016single}, let
$R={(i_k,j_k,r_k)},k=1\dots K$ be the set of relative depth annotations,
where $i_k$ and $j_k$ are the locations of two points in the $k$-th annotation and $r_k\in
\{>,<,=\}$ is the  ground-truth ordinal relation (closer, further, or same distance). 
Let $S=\{p_l, n_l\}$ be the set of surface normal annotations, where $p_l$ is the location
of the $l$-th annotation and $n_l \in \mathbf{R}^3$ is the ground truth surface normal at
this location. 

We can now express the loss function as follows: 
\begin{equation}
L(R, S, z) = \frac{1}{K} \sum_{k=1}^K \psi(i_k,j_k,r_k, z) + \lambda \frac{1}{L} \sum_{l=1}^L \phi(p_l,n_l, z)
\label{equation:relative_depth_plus_normal}
\end{equation}
where $z$ is the depth map predicted by the network. The loss term $\psi(i_k,j_k,r_k, z)$
measures the inconsistency between the predicted depth map $z$ and the $k$-th relative depth
annotation. The loss term $\sum_{l=1}^L \phi(p_l,n_l, z)$ measures the inconsistency
between the predicted depth map $z$ and the $l$-th surface normal annotation. The
hyper-parameter $\lambda$ balances the two terms.

\noindent
\textbf{A revised relative depth loss}
Chen et al.~\cite{chen2016single} define the loss term $\psi(i_k,j_k,r_k, z)$ as 
\begin{align}
\left\{
\begin{array}{ll}
\ln\left(1+\exp(-z_{i_k} + z_{j_k}) \right), & r_k \in \{>\} \\
\ln\left(1+\exp(z_{i_k} - z_{j_k}) \right),  & r_k \in \{<\} \\
(z_{i_k} - z_{j_k})^2, &  r_k \in \{=\} \\
\end{array}
\right.
\label{equation:definition_of_old_relative_loss}
\end{align}

This definition encourages two depth values to be as different as possible if
their ground truth ordinal relation is an inequality, or as similar as possible if their ground
truth relation is equality. It works well if relative depth is the only form of
supervision, as shown by Chen et al.~\cite{chen2016single}, but it is problematic when
used in conjunction with annotations of surface normals. The
problem is that it encourages the difference of two unequal depth values to be infinitely
large. This can potentially conflict with annotations of surface normals, which encourage
the depth values to have a specific difference to form a specific surface orientation.

To address this issue we revise the loss term by introducing a margin $\tau > 0$ that
stops the loss from decreasing if two depth values supposed to be unequal are already at least $\tau$
apart and if two equal depth values supposed to be equal are apart by no more than
$\tau$: 
 \begin{align}
 \left\{
 \begin{array}{ll}
 \ln\left(1+\exp(- \min( z_{i_k} - z_{j_k}, \tau ) \right)), & r_k \in \{>\} \\
 \ln\left(1+\exp(- \min( z_{j_k} - z_{i_k}, \tau ) \right)),  & r_k \in \{<\} \\
 \max(\tau^2,|z_{i_k} - z_{j_k}|^2), &  r_k \in \{=\}. \\
 \end{array}
 \right.
 \label{equation:definition_of_new_relative_loss}
 \end{align}

To make the loss term compatible with surface normals, we make another modification. We add
a softplus transform to the network to enforce positive depth. This is needed because
a negative depth means that the object is behind the camera and will cause issues in computing
surface normals from the predicted depth. 
 
\smallskip
\smallpara{Angle-based surface normal loss}
We now consider how to define the loss term $\phi(p_l,n_l, z)$ in
Eqn.~\ref{equation:relative_depth_plus_normal} that compares the predicted depth map $z$
 with a ground truth surface normal $n_l$ at location $p_l$. 

The first approach we propose is to derive a surface normal $\nu(z)_{p_l}$ at the same location from the
predicted depth map $z$ and compare the derived normal to the ground truth. Here $\nu$ is
a function that maps a depth map to a map of surface normals, and $\nu(z)_{p_l}$ is the
derived surface normal at location $p_l$. The loss term can now be defined as the
angular difference between the derived normal and the ground truth normal, expressed as a
dot product of the two normals: 
\begin{equation}
\begin{split}
\phi_l(p_l, n_l, z) = - <n_l, \nu(z)_{p_l}>. 
\end{split}
\label{eqn:angle_loss}
\end{equation}
We call this formulation the \emph{angle-based surface normal loss}.

To derive surface normals from depth, i.e.\@ to implement the function $\nu$, we first back-project the pixels to 3D points in the
camera coordinate system, assuming a pinhole camera model with a known focal length
$f$. In particular, a pixel located at $(x,y)$ on the image plane with depth $z'$ is mapped to the 3D point $(xz'/f, yz'/f,
z')$: 
\begin{equation}
\beta: (x,y,z') \rightarrow (xz'/f, yz'/f, z')
\label{eqn:back-proj}
\end{equation}
We then compute the surface normal $\nu(z)_{xy}$ for a pixel located at $(x,y)$ 
using the cross product of the two vectors formed by its adjacent four
neighbors (top to bottom, left to right):
\begin{equation}
\begin{split}
\nu(z)_{xy} =  [\beta(x-1,y,z_{x-1,y}) - \beta(x+1,y,z_{x+1,y})] \\ 
		 \otimes [\beta(x,y-1,z_{x,y-1}) - \beta(x,y+1,z_{x,y+1})],
\end{split}
\label{equation:normal_cross_prod}
\end{equation}
where $\otimes$ denotes cross product and $\beta$ is the back-projection function in
Eqn.~\ref{eqn:back-proj}. Combining Eqn.~\ref{eqn:back-proj}, and
Eqn.~\ref{eqn:angle_loss} gives a loss term $\phi(p_l,n_l, z)$ that is differentiable with respect to the
predicted depth $z$ and can be easily incorporated into backpropagation. 

\smallskip
\smallpara{Depth-based surface normal loss.} The angle-based surface normal loss is
natural, and a network trained with this loss in addition to relative depth annotations should predict better depth, as measured by the
metric error (comparing the predict depth with ground truth depth in terms of absolute
difference). 
In our experiments, however, we observe that this is not always the case,
especially with a large training set. In particular, we observe that a network will
predict a depth map that gives better surface normals, but the depth map itself does not improve
 in terms of metric error. 

\begin{figure}[t]
	\begin{center}		
		\includegraphics[width=\linewidth]{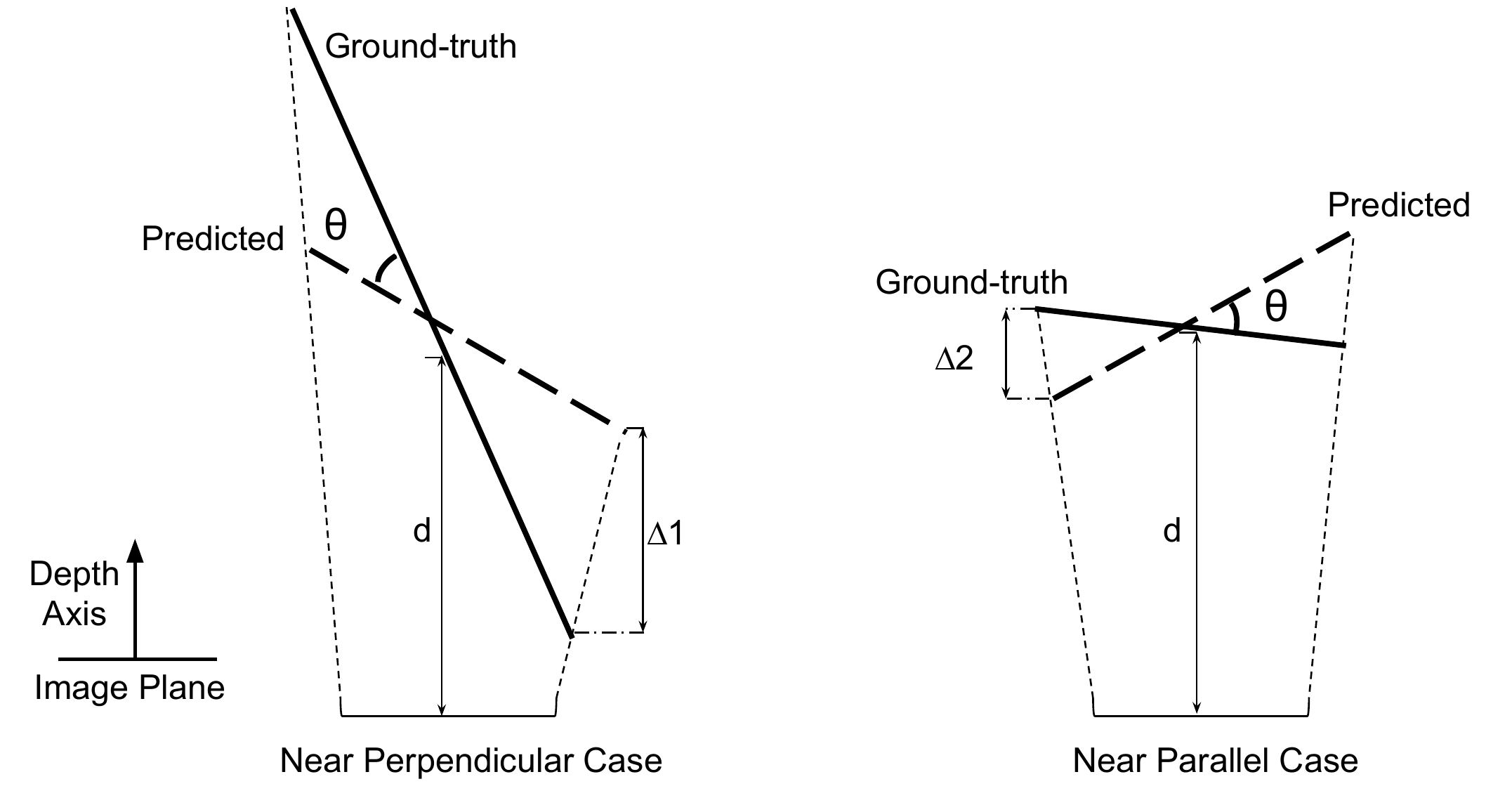}
	\end{center}
	\vspace{-3mm}
	\caption{Two 3D planes (solid line) whose centers have the same distance $d$ to the image 
	plane and whose projections occupy the same amount of area on an image. The predicted
        surface normals both deviate by $\theta$ from the ground-truth, but incur drastically 
	different metric depth errors $\Delta1$ and $\Delta2$.}
	\label{fig:small_normal_diff_large_metric_error}
\end{figure}

This leads us to make one theoretical observation. The observation is that
 when a surface normal is pointing sideways, a small change of
the surface normal corresponds to a disproportionally large change in depth values for the
neighobring pixels. In other words, metric depth error is very sensitive to the depth
values in regions of steep slopes, but the angle-based loss does not reflect this
sensitivity (Fig.~\ref{fig:small_normal_diff_large_metric_error}). 
This could result in the phenomenon that a decrease in the angle-based loss
does not corresponds to any notable improvement of metric depth error---the network is not focusing
on the steep slopes, the places that would make the most difference in metric depth error. 

Based on this observation we propose an alternative loss formulation, which we call
\emph{depth-based surface normal loss}.  The idea is to take the predicted depth at a
pixel and compute depth value of a neighbor using the ground truth normal. In other words,
we compute the depth value the neighbor should take in order to be fully consistent with the ground
truth normal. This ``should-be'' depth is compared with the actual predicted depth for the
neighbor, and the difference becomes the penalty in the loss term. This loss is
essentially converting a surface normal into the derivative of depth, and then compare it
to the actual predicted derivative of depth. This
depth-based loss is thus better aligned with metric depth error: surface normal annotations at
steep slopes will play a bigger role in the loss. 

Specifically, let $p^{T}$, $p^{B}$, $p^{L}$, $p^{R}$ be the top, bottom, left, right neighbors of pixel $p$. We first
obtain the back projection $X^T$ of $p^{T}$ using the predicted depth $z_{p^T}$ (same as in Eqn.~\ref{eqn:back-proj}). Let $\Pi^T$
denote the plane that goes through $X^T$ and is oriented according to the ground truth
normal $n_p$. By intersecting $\Pi^T$ with a ray that originates from the camera center
and goes through the bottom neighbor $p^{T}$ in the
image plane, we obtain the ``should-be'' depth value $\hat{z}_{p^B}$ for the bottom neighbor $p^{B}$. Similarly, we can
obtain the ``should-be'' depth value for the top neighbor from the bottom neighbor
($\hat{z}_{p^T}$ from $z_{p^B}$), for the left neighbor from the right neighbor ($\hat{z}_{p^L}$ from
$z_{p^R}$), and for the right neighbor from the left neighbor ($\hat{z}_{p^R}$ from $z_{p^L}$). 
Finally, the loss term is defined as the difference between
the ``should-be'' depth and the actual predicted depth for all neighbors. 
\begin{equation}
\phi_l(p_l, n_l, z)  = \sum_{i\in\{T,B,L,R\} }(\hat{z}_{p_l^i} - z_{p_l})^2 / (\hat{z}_{p_l^i} + z_{p_l})^2,
\end{equation}
which is differentiable with respect to $z$. Note that the squared difference between the
two depth values is normalized by their squared sum. This is for scale invariance;
otherwise the network will minimize the loss mostly by shrinking the depth
values with little regard to the normals. 

\smallskip
\smallpara{Multiscale normals} In addition to introducing depth-based loss, we consider yet
another strategy to address the issue of angle-based surface normal loss. The strategy is
to collect surface normal annotations at multiple resolutions. That is, we can collect some surface
normal annotations at lower resolutions. The
rationale is that the steep slopes get smoothed out in lower resolutions and become less
steep, which brings the angle-based loss more in line with metric depth error. To use
the normals from lower resolutions, we add downsampling layers to the network to produce
depth maps of lower resolutions, and add an angle-based loss at each additional resolution of
the depth map.

\section{Experiments on NYU Depth}

We perform extensive experiments on NYU Depth~\cite{silberman2012indoor}. The ground truth
metric depth available in NYU Depth allows us to simulate and evaluate how adding surface normal
annotations as indirect supervision can improve the prediction of metric depth, which is
impossible for images in the wild, which do not have metric depth ground truth. 

\smallpara{Implementation details}
For all our experiments on NYU Depth, we use the same network architecture
proposed in ~\cite{chen2016single}. The only difference is two modifications made to ensure that the loss
term on relative depth will not encourage the predicted depth to deviate from the true metric depth, thus minimizing
conflict with the loss term on surface normals.  First, we add a softplus
layer to ensure positive depth. Second, we take the log of the predicted depth before
sending it to the relative depth loss in Eqn.~\ref{equation:definition_of_new_relative_loss}. Taking the difference of the log depth is the same as taking the log of the depth
ratio, which is more consistent with the relative
depth annotations in NYU Depth~\cite{chen2016single,zoran2015learning} because the ground
truth ordinal
depth relations are based on thresholding depth ratios rather than thresholding depth difference.

For relative depth ``annotations'' on NYU Depth, we use the same set as in
~\cite{chen2016single}. For surface normal ``annotations'', we generate them from the ground-truth
depth using Eq~\ref{equation:normal_cross_prod}. Unless otherwise
noted, in all our models trained with surface normals, we provide 5,000 surface normal annotations at 
random locations per image. 

\smallskip
\smallpara{Main experiments}
We compare 5 models: \textbf{(1)} a model trained with relative depth only (\textit{d}); \textbf{(2)} a model
trained with relative depth and surface normals using the angle-based loss
(\textit{d\_n\_al});  
\textbf{(3)} same as (2) but using surface normals from multiple resolutions while keeping
the total number of normal samples the same (\textit{d\_n\_al\_M}). 
\textbf{(4)} a model trained with relative depth and surface normals using depth-based loss (\textit{d\_n\_dl}). 
\textbf{(5)} same as (4) but using surface normals from multiple resolutions while keeping 
the total number the same (\textit{d\_n\_dl\_M}). 

As in prior work~\cite{chen2016single,zoran2015learning}, for each of the 5 models we train and evaluate on \emph{NYU Subset}, a standard subset of
1449 images in NYU Depth, and \emph{NYU Full}, the entire NYU Depth. Models
trained on NYU Full are named with a \textit{\_F} suffix).  In this section we discuss quantitative results.
For qualitative results, please refer to the Appendix.

\smallskip
\smallpara{Evaluating metric depth}
Metric depth error measures the metric differences between the predicted depth map
and the ground-truth depth map. Following prior
work~\cite{chen2016single,eigen2015predicting,zoran2015learning}, we evaluate the root mean squared error
(RMSE), the log RMSE, the log scale-invariant RMSE (log RMSE(s.inv)), the absolute
relative difference (absrel) and the squared relative difference (sqrrel); their precise definitions
can be found in~\cite{eigen2014depth}. Because single-image depth has scale ambiguity,
before evaluation we normalize each predicted depth map such that it has the same mean and
variance as those of the entire training set, as is done in~\cite{chen2016single}. 

However, such normalization is too crude in that it forces
every predicted depth map to have the same mean and variance regardless of the input
 scene, which will unfairly penalize accurate predictions for scenes with a different
 mean and variance. We therefore propose a new error metric \emph{Least-Square 
 RMSE} (LS-RMSE) that better handles scale ambiguity in evaluation: for a predicted depth map $z$ and
its ground-truth $z^*$ with pixels indexed by i, we compute the smallest possible sum of their squared
differences under a global scaling and translation of the depth values: 
\begin{equation}
\mathrm{LS\_RMSE}(z, z^*) = \min_{a,b} \sum_i (az_i + b - z^*_i)^2.
\end{equation}
Note that computing this error metric is the same as finding the least square solution to
a system of linear equations, which has a well-known closed form solution.

Tab.~\ref{table:metric_error} reports the results on metric depth error. We can see that
our baseline model trained with relatived depth only matches or exceeds the metric depth error
reported by Chen et al.~\cite{chen2016single}. We attribute this improvement to our
revised relative depth loss (Eqn.~\ref{equation:definition_of_new_relative_loss}), which does not encourage exaggerating depth differences once
the ordering is correct. 

On both NYU Subset and NYU Full, adding surface normals in training achieves significant improvement in metric depth
quality, as reflected most notably in LS-RMSE. The improvement in metrics other than LS-RMSE is less significant, indicating a mismatch
 of depth scale. Among the models trained with surface normals, the one
trained with the depth-based loss (\textit{d\_n\_dl\_F}) performs the best, as expected from our discussion in
Sec.~\ref{section:Learning_with_surface_normals}. On NYU Full, it outperforms the relative-depth-only baseline significantly on
 \textit{LS RMSE}, approaching the models trained with full ground truth
metric depth maps (Eigen(V)~\cite{eigen2015predicting}, Chakrabarti~\cite{chakrabarti2016depth}). 

The model trained with the angle-based 
normal loss yields no improvemet on NYU Subset and negative improvement on NYU Full, which
can be explained by our theoretical observation that the angle-based loss is misaligned
with the metric depth error. The misalignment is especially notable on a bigger dataset,
which is harder to fit and can cause the network to ``give up'' on the steep slopes, which
account for very little in the angle-based normal loss. Using multiscale normals 
helps as expected, but it is not enough to overcome the misalignment on NYU Full to outperform the
relative-depth-only baseline.

\smallskip
\smallpara{Evaluating relative depth}
We also evaluate a predicted depth map on ordinal error: disagreement with ground truth
ordinal relations between selected locations. We use the same set of ground truth ordinal
relations from ~\cite{chen2016single}, and report the same metrics: WKDR, the weighted 
disagreement rate between the
predicted ordinal relations and the ground-truth ordinal relations, and its variants
WKDR$^{=}$ (WKDR of pairs whose ground-truth order is =) and WKDR$^{\neq}$(WKDR of pairs
whose ground-truth order is either $>$ or $<$). 
 
Following ~\cite{chen2016single}, we predict the ordinal relation of point A and B
by thresholding on difference of the predicted depth. 

The results on relative depth are shown in 
Tab.~\ref{table:relative_depth_performance}. First it is interesting to observe that our relative-depth-only
baseline model is slightly worse than Chen et al.~\cite{chen2016single}, which also trains with
only relative depth. We attribute this difference to our revised relative depth loss
(Eqn.~\ref{equation:definition_of_new_relative_loss})---the loss in Chen et
al.~\cite{chen2016single} encourages exaggerating depth differences, which leads to better
relative depth performance at the expense of metric accuracy, as reflected by
Tab.~\ref{table:metric_error}.

Interestingly, adding normals improves ordinal error, but only from the
angle-based normal loss, not from the depth-based normal loss. This is because depth-based
normal loss places great emphasis on getting the exact steep slopes, but this does not
make any difference to ordinal error as long as the sign of the slope is
correct.

\begin{table}
\begin{center}
\resizebox{\columnwidth}{!}{%
\begin{tabular}{|c| l | c | c | c | c | c | c|}
\hline
Training	&	Method & RMSE     & RMSE     & log RMSE & absrel   & sqrrel  & LS  \\
Data		&		   &      	  & (log)    &(s.inv)   &          &         & RMSE         \\
\hline
NYU			& d  	   &	1.12	& 0.39	& 0.26	& 0.36	& 0.45	&  	 0.64\\		
Subset		& d\_n\_al &	1.13	& 0.39  & 0.26	& 0.36	& 0.45	&	 0.65\\
			& d\_n\_al\_M &		1.11	&	0.39	& 0.25	&	0.36	& 0.44	&	0.59 \\
			& \textbf{d\_n\_dl} &	\textbf{1.11}	& \textbf{0.39}	& \textbf{0.25}	& \textbf{0.35}	& \textbf{0.44}	&	 \textbf{0.58}\\
			& d\_n\_dl\_M &	1.11	& 0.39	& 0.25	& 0.36	& 0.45	&	 0.59 \\ 
			& Chen~\cite{chen2016single}     & 1.12	& 0.39 & 0.26	& 0.36	& 0.46	&	 0.65  \\
			& Zoran~\cite{zoran2015learning} & 1.20 & 0.42 & - & 0.40  & 0.54 &   -\\
\hline
	NYU		& d\_F 		  & 1.08	&	0.37	& 	0.23	&	0.34	&	0.41	&	0.52 \\ 
	Full	& d\_n\_al\_F &	1.09	&	0.38	&	0.24	&	0.34	&	0.42	&	0.55 \\
			& d\_n\_al\_F\_M &	1.09 &	0.38	&	0.23	&	0.34	&	0.41	&	0.53 \\
			& \textbf{d\_n\_dl\_F} & \textbf{1.08} &	\textbf{0.37} & \textbf{0.23} & \textbf{0.34} & \textbf{0.41} &	 \textbf{0.50} \\ 	
			& d\_n\_dl\_F\_M & 1.09 &	0.38 & 0.24 & 0.35 & 0.43 &	0.52  \\ 	
			& Chen\_Full~\cite{chen2016single} & 1.09 & 0.38 & 0.24 & 0.34 & 0.42 &  0.58\\	\cline{2-8}		
		    & Eigen(V)*~\cite{eigen2015predicting} & 0.64 & 0.21 & 0.17 & 0.16 & 0.12 &   0.47\\ 
		    & Chakrabarti*~\cite{chakrabarti2016depth} & 0.64& 0.21& 0.17& 0.15 & 0.12 & 0.47\\	

\hline
\end{tabular}
}
\end{center}
\caption{ Metric depth error evaluated on the NYU Depth dataset. Models with a * suffix are trained on full metric depth.}
 \label{table:metric_error}
\end{table}

\begin{table}
\begin{center}
\resizebox{0.8\columnwidth}{!}{%
\begin{tabular}{|c | l | c | c | c|}
\hline
	Training &	Method & WKDR & $\text{WKDR}^{=}$ & $\text{WKDR}^{\neq}$ \\ 
	Data & & & & \\
\hline
	NYU		&		 			d		 &  37.6\%  &  36.4\%  & 39.3\% \\ 	  
	Subset	&		 			d\_n\_al &  36.5\%  &  35.5\%  & 37.9\% \\   
			&		 		 \textbf{d\_n\_al\_M} &  \textbf{34.6\%}  &  \textbf{33.4\%}  & \textbf{36.3\%} \\   
			&	  			 	d\_n\_dl &  38.7\%  & 36.9\%   & 40.5\%  \\   
			&	  			 d\_n\_dl\_M &  39.0\%  & 37.7\%   & 40.5\%  \\   
			& Chen~\cite{chen2016single} &  35.6\%  & 36.1\%   & 36.5\% \\
			& Zoran~\cite{zoran2015learning}	&  43.5\%  &  44.2\%  & 41.4\% \\

\hline
NYU			&		   	d\_F    	   & 29.2\%&  32.5\%& 28.0\%\\  
Full		&		   	\textbf{d\_n\_al\_F}    & \textbf{27.6\%} & \textbf{ 31.5\%} & \textbf{26.6\%}\\  % 
			&		   	d\_n\_al\_F\_M & 27.9\% &  32.2\% & 26.6\%\\  
			&		   	d\_n\_dl\_F    & 30.9\%  & 31.7\% & 31.4\%\\  
			&	  	    d\_n\_dl\_F\_M & 35.5\%  & 38.9\% & 34.6\%  \\   
		    & Chen\_Full~\cite{chen2016single} & 28.3\%  & 30.6\% & 28.6\% \\  \cline{2-5}

   			& Eigen(V)*~\cite{eigen2015predicting} &  34.0\%  & 43.3\%   & 29.6\% \\
			& Chakrabarti*~\cite{chakrabarti2016depth} & 27.5\%& 30.0\%& 27.5\%\\
\hline
\end{tabular}
}
\end{center}
\caption{Ordinal error evaluated on the NYU Depth dataset. Models with a * suffix are trained on full metric depth.}
 \label{table:relative_depth_performance}
\end{table}

\begin{figure*}[t]
	\begin{center}
		\includegraphics[width=\linewidth]{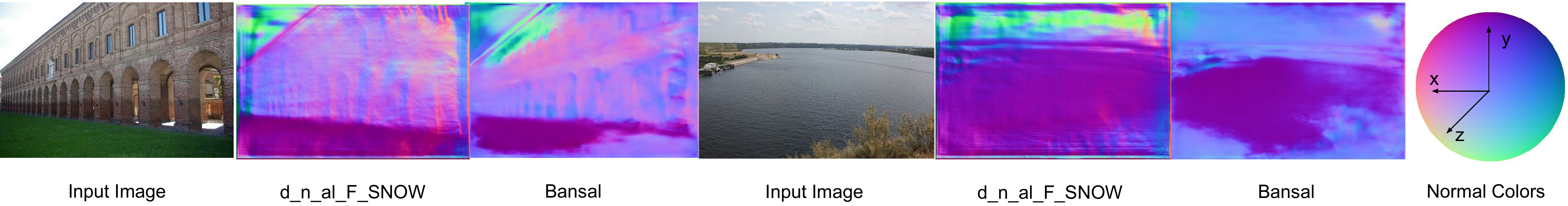}
	\end{center}	
	\caption{Normal maps produced by our model and Bansal~\cite{bansal2016marr}. Please view in color. More examples are in the Appendix.}
	\label{fig:qual_result_snow}
\end{figure*}

\smallskip
\smallpara{Evaluating surface normals}
We now evaluate the predicted depth in terms of surface normals derived from it. 
We use the same metrics as in~\cite{eigen2015predicting}: the mean and median of angular difference with
the ground-truth, and the percentages of predicted samples whose angular difference with
the ground-truth are under a certain threshold. The ground truth normals for test
are from NYU Depth toolkit~\cite{silberman2012indoor}, as is done
in~\cite{wang2015designing,eigen2015predicting}. We also evaluate the \emph{derived} surface
normals from other depth-estimation models, including \textbf{(1)} state-of-the-art depth estimation method of
Eigen~\cite{eigen2015predicting} and Chakrabarti~\cite{chakrabarti2016depth}; \textbf{(2)} 
The original method of Chen et al.~\cite{chen2016single} augmented with a softplus layer
to ensure positive depth but otherwise trained the same way with relative depth only
(Chen* and Chen\_Full*). 

We report the results in Tab.~\ref{table:normal_error}. As expected, models trained with
the angle-based normal loss perform better
than any other models in terms of surface normals derived from depth, as the loss
directly targes the normal error metric. 

For reference, we also evaluate state of art methods that \emph{directly predict} surface normals:
Bansal~\cite{bansal2016marr}, Eigen~\cite{eigen2015predicting}
and Wang~\cite{wang2015designing}. Note that these models
are trained on the full dense normal maps on NYU Full whereas our models are trained with only a sparse set of
normals. Yet our best model (\textit{d\_n\_al\_F}) outperforms Wang~\cite{wang2015designing}.

\begin{table}
\begin{center}
\resizebox{\columnwidth}{!}{%
\begin{tabular}{| c | l | c   c | c  c  c |}
\hline
		Training	&	 	Method	& \multicolumn{2}{c|}{Angle Distance}& \multicolumn{3}{c|}{\% Within $t^{\circ}$}        \\
		Data 		&  		 		& 			Mean  &		Median 		 & 11.25$^{\circ}$ & 22.5$^{\circ}$ & 30$^{\circ}$\\
\hline
\hline
NYU		 & 	d			& 45.46& 40.62&	 7.56&	23.65& 	35.10\\ 
Subset		 & 	d\_n\_al	& 37.53& 31.93&	13.04&	34.38& 	47.39\\
		 & 	\textbf{d\_n\_al\_M}	& \textbf{35.39}& \textbf{29.51}&	\textbf{15.50}&	\textbf{38.43}& 	\textbf{51.40}\\ 
		 & 	d\_n\_dl	& 40.53& 34.58&	11.40&	31.13& 	43.56\\ 
		 & 	d\_n\_dl\_M	& 41.88& 35.76&	10.73&	29.69& 	41.88\\ 
		 & Chen*~\cite{chen2016single}	& 50.68&	44.96&	4.16&	16.77&	28.21	\\   
		 
\hline
NYU	   &   			 	d\_F	  & 29.45	& 22.71	& 22.31& 50.71&	63.65\\  	  
Full   &   			 	\textbf{d\_n\_al\_F}	  & \textbf{25.92}	& \textbf{20.09}	& \textbf{26.28}& \textbf{56.45}&	\textbf{69.26}\\  	  
	   &   			 	d\_n\_al\_F\_M& 26.50 & 20.42 & 26.41& 55.47& 68.09	\\  	  
	   &   			 	d\_n\_dl\_F	& 30.85 & 24.51	& 24.51	& 46.93& 60.31	\\  	  
	   & 			 d\_n\_dl\_F\_M	& 37.63& 31.58&	13.41&	34.97& 47.97	\\ 
	   & Chen\_Full*~\cite{chen2016single}& 30.35&	24.37&	18.64&	46.80&	61.42	\\ 
  	   & Eigen(V)~\cite{eigen2015predicting}& 35.97 &   28.34  & 17.67  & 41.12  &  53.49 	  \\  
	   & Chakrabarti~\cite{chakrabarti2016depth}& 29.80  &  20.43   & 31.34  & 54.90  &  64.57 	  \\ \cline{2-7}  	   
	   & Wang\S~\cite{wang2015designing}& 28.8 &     17.9&  35.2 & 57.1  &  65.5 	  \\
       & Eigen(V)\S~\cite{eigen2015predicting}&  22.89&  16.26&  38.23&   63.30& 73.18  	  \\   
 	   & Bansal\S~\cite{bansal2016marr}&  	\textbf{22.63}& \textbf{15.78} &   \textbf{39.17} &  \textbf{64.17}&  \textbf{73.77} 	  \\  
\hline
\end{tabular}
}
\end{center}
\caption{Surface normals error evaluated on the NYU Depth dataset. The lower the better for Angle Distance metrics. The higher the better for the Percentage within $t^{\circ}$ metrics. Models with a \S~suffix directly predict surface normals.  }
 \label{table:normal_error}
\end{table}
  
\smallpara{Discussion} Our expriements on NYU Depth show that surface normal annotations
can help depth estimation in the absence of ground truth depth.  We have proposed
two different surface normal losses. Each has a different set of trade-offs and is
appropriate in different applications. If metric fidelity is important, especially at depth discontinuities, then the depth-based loss is more
appropriate. If surface orientation is important than the fidelity of depth discontinuities, then the angle-based loss is more
appropriate.

\begin{table}
	\begin{center}
		\resizebox{\columnwidth}{!}{%
			\begin{tabular}{| c | l | c   c | c  c  c |}
				\hline
         &						Model	    & \multicolumn{2}{c|}{Angle Distance}& \multicolumn{3}{c|}{Within $t^{\circ}$}        \\
    	 &								    & Mean &  Median & 11.25$^{\circ}$ & 22.5$^{\circ}$ & 30$^{\circ}$\\
						\hline
						\hline				
Normals  &			    d\_n\_al\_F	&   32.53&      27.44&      15.40&  40.52&  54.12	 	  \\ 
from	 &	\textbf{d\_n\_al\_F\_SNOW}	& \textbf{25.75}&      \textbf{21.26}&      \textbf{21.66}&  \textbf{52.98}&  \textbf{67.88}		  \\ 
Predicted&	Chen\_Full~\cite{chen2016single}& 35.16&	30.26&	13.70&	36.56&	49.56 	  \\ 
Depth    & Eigen(V)~\cite{eigen2015predicting}& 48.71&	46.15&	6.35&	18.91&	28.45	  \\
	     &		 FCRN~\cite{laina2016deeper}& 48.74&	45.38&	5.84&	18.29&	28.25 	  \\
				\hline	
Directly &			Ours\_NYU\S      & 31.96  &	  26.03&	  18.16&	  43.72&	  56.03	  \\	
Predicted&		  \textbf{Ours\_NYU\_SNOW\S}	 & \textbf{23.33}  &	  \textbf{17.99}&	  \textbf{30.42} &	  \textbf{60.54}&	  \textbf{72.74}	  \\ 
Normals  & Eigen(V)\S~\cite{eigen2015predicting}& 28.71&	23.16&	20.98&	48.78&	61.84	 	  \\
	     &		Bansal\S~\cite{bansal2016marr}& 27.85&	22.25&	23.41&	50.54&	64.09		 \\
						\hline
			\end{tabular}
		}
	\end{center}
	\caption{Surface normals error evaluated on SNOW. Models with a \S~suffix directly predict surface normals. }
	\label{table:snow_performance}
\end{table}

\section{Experiments on SNOW}

Since SNOW provides no ground truth of metric depth, 
it is infeasible to evaluate how training with surface normals helps predict
metric depth. We thus evaluate surface normals as an indirect indicator of depth quality
for images in the wild.  We split SNOW into 10,256 test images and 49,805 training images. 

We first evaluate the surface normals \emph{derived} from depth prediction. 
Our baselines include state-of-the-art depth estimation methods
Eigen~\cite{eigen2015predicting} and FCRN~\cite{laina2016deeper}, 
both trained with full metric depth from NYU Full. 
We compare these baselines with the \textit{d\_n\_al\_F} network, our best performing
model in terms of normal error. We also fine tune the \textit{d\_n\_al\_F}
network on SNOW (\textit{d\_n\_al\_F\_SNOW}).

We can see in Tab.~\ref{table:snow_performance} that our network 
trained only on NYU Full (\textit{d\_n\_al\_F}) already outperforms the baselines. Fine-tuning
on SNOW yields a significant improvement. 

SNOW also enables us to evaluate on methods that \emph{directly predict} surface normals. We include
 four models: \textbf{(1)} state-of-the-art surface normal 
estimation methods of Bansal~\cite{bansal2016marr} and Eigen~\cite{eigen2015predicting}; \textbf{(2)} Chen et al.~\cite{chen2016single}'s network 
trained to directly predict normals (\textit{Ours\_NYU\S}); \textbf{(3)} \textit{Ours\_NYU\S} fine-tuned on SNOW (\textit{Ours\_NYU\_SNOW\S}). 
We can see from Tab.~\ref{table:snow_performance} that fine-tuning on SNOW significantly
improves surface normal prediction. Finally, Fig.~\ref{fig:qual_result_snow} shows
examples of qualitative improvement achieved by our network on images in the wild.

\section{Conclusion} We have proposed two distinct approaches for using surface 
normal annotations to train a deep network that directly predicts per-pixel metric depth. 
We have also introduced a new dataset of crowdsourced surface normals for images in the wild 
(SNOW). Experiments show that surface normal annotations can advance depth estimation in the wild.

\medskip
\smallpara{Acknowledgments}
This work is partially supported by the National Science Foundation under Grant No. 1617767.

{\small
\bibliographystyle{ieee}
\bibliography{egbib}

\begin{thebibliography}{10}\itemsep=-1pt

\bibitem{baig2015coupled}
M.~H. Baig and L.~Torresani.
\newblock Coupled depth learning.
\newblock {\em arXiv preprint arXiv:1501.04537}, 2015.

\bibitem{bansal2016marr}
A.~Bansal, B.~Russell, and A.~Gupta.
\newblock Marr revisited: 2d-3d alignment via surface normal prediction.
\newblock In {\em Proceedings of the IEEE Conference on Computer Vision and
  Pattern Recognition}, pages 5965--5974, 2016.

\bibitem{BarronTPAMI2015}
J.~T. Barron and J.~Malik.
\newblock Shape, illumination, and reflectance from shading.
\newblock {\em TPAMI}, 2015.

\bibitem{bell13opensurfaces}
S.~Bell, P.~Upchurch, N.~Snavely, and K.~Bala.
\newblock Open{S}urfaces: A richly annotated catalog of surface appearance.
\newblock {\em ACM Trans. on Graphics (SIGGRAPH)}, 32(4), 2013.

\bibitem{Butler:ECCV:2012}
D.~J. Butler, J.~Wulff, G.~B. Stanley, and M.~J. Black.
\newblock A naturalistic open source movie for optical flow evaluation.
\newblock In {\em ECCV}, Part IV, LNCS 7577, pages 611--625. Springer-Verlag,
  Oct. 2012.

\bibitem{chakrabarti2016depth}
A.~Chakrabarti, J.~Shao, and G.~Shakhnarovich.
\newblock Depth from a single image by harmonizing overcomplete local network
  predictions.
\newblock In {\em Advances in Neural Information Processing Systems}, pages
  2658--2666, 2016.

\bibitem{chen2016single}
W.~Chen, Z.~Fu, D.~Yang, and J.~Deng.
\newblock Single-image depth perception in the wild.
\newblock In {\em Advances in Neural Information Processing Systems}, pages
  730--738, 2016.

\bibitem{dai2017scannet}
A.~Dai, A.~X. Chang, M.~Savva, M.~Halber, T.~Funkhouser, and M.~Nie{\ss}ner.
\newblock Scannet: Richly-annotated 3d reconstructions of indoor scenes.
\newblock {\em arXiv preprint arXiv:1702.04405}, 2017.

\bibitem{eigen2015predicting}
D.~Eigen and R.~Fergus.
\newblock Predicting depth, surface normals and semantic labels with a common
  multi-scale convolutional architecture.
\newblock In {\em ICCV}, 2015.

\bibitem{eigen2014depth}
D.~Eigen, C.~Puhrsch, and R.~Fergus.
\newblock Depth map prediction from a single image using a multi-scale deep
  network.
\newblock In {\em NIPS}, 2014.

\bibitem{gallianijust}
S.~Galliani and K.~Schindler.
\newblock Just look at the image: viewpoint-specific surface normal prediction
  for improved multi-view reconstruction.
\newblock 2016.

\bibitem{garg2016unsupervised}
R.~Garg, G.~Carneiro, and I.~Reid.
\newblock Unsupervised cnn for single view depth estimation: Geometry to the
  rescue.
\newblock In {\em European Conference on Computer Vision}, pages 740--756.
  Springer, 2016.

\bibitem{geiger2013vision}
A.~Geiger, P.~Lenz, C.~Stiller, and R.~Urtasun.
\newblock Vision meets robotics: The kitti dataset.
\newblock {\em The International Journal of Robotics Research}, page
  0278364913491297, 2013.

\bibitem{hane2015direction}
C.~Hane, L.~Ladicky, and M.~Pollefeys.
\newblock Direction matters: Depth estimation with a surface normal classifier.
\newblock In {\em Proceedings of the IEEE Conference on Computer Vision and
  Pattern Recognition}, pages 381--389, 2015.

\bibitem{ikehata2016panoramic}
S.~Ikehata, I.~Boyadzhiev, Q.~Shan, and Y.~Furukawa.
\newblock Panoramic structure from motion via geometric relationship detection.
\newblock {\em arXiv preprint arXiv:1612.01256}, 2016.

\bibitem{Karsch:TPAMI:14}
K.~Karsch, C.~Liu, and S.~B. Kang.
\newblock Depthtransfer: Depth extraction from video using non-parametric
  sampling.
\newblock {\em TPAMI}, 2014.

\bibitem{Kazhdan:2006}
M.~Kazhdan, M.~Bolitho, and H.~Hoppe.
\newblock Poisson surface reconstruction.
\newblock SGP. Eurographics Association, 2006.

\bibitem{koenderink1992surface}
J.~J. Koenderink, A.~J. Van~Doorn, and A.~M. Kappers.
\newblock Surface perception in pictures.
\newblock {\em Attention, Perception, \&amp; Psychophysics}, 52(5):487--496,
  1992.

\bibitem{kushal2013single}
A.~Kushal and S.~M. Seitz.
\newblock Single view reconstruction of piecewise swept surfaces.
\newblock In {\em 2013 International Conference on 3D Vision-3DV 2013}, pages
  239--246. IEEE, 2013.

\bibitem{ladicky2014pulling}
L.~Ladicky, J.~Shi, and M.~Pollefeys.
\newblock Pulling things out of perspective.
\newblock In {\em CVPR}, 2014.

\bibitem{laina2016deeper}
I.~Laina, C.~Rupprecht, V.~Belagiannis, F.~Tombari, and N.~Navab.
\newblock Deeper depth prediction with fully convolutional residual networks.
\newblock In {\em 3D Vision (3DV), 2016 Fourth International Conference on},
  pages 239--248. IEEE, 2016.

\bibitem{li2015depth}
B.~Li, C.~Shen, Y.~Dai, A.~van~den Hengel, and M.~He.
\newblock Depth and surface normal estimation from monocular images using
  regression on deep features and hierarchical crfs.
\newblock In {\em CVPR}, 2015.

\bibitem{liu2015deep}
F.~Liu, C.~Shen, and G.~Lin.
\newblock Deep convolutional neural fields for depth estimation from a single
  image.
\newblock In {\em CVPR}, 2015.

\bibitem{ranftldense}
R.~Ranftl, V.~Vineet, Q.~Chen, and V.~Koltun.
\newblock Dense monocular depth estimation in complex dynamic scenes.
\newblock In {\em CVPR}, 2016.

\bibitem{richter2016playing}
S.~R. Richter, V.~Vineet, S.~Roth, and V.~Koltun.
\newblock Playing for data: Ground truth from computer games.
\newblock In {\em European Conference on Computer Vision}, pages 102--118.
  Springer, 2016.

\bibitem{roymonocular}
A.~Roy and S.~Todorovic.
\newblock Monocular depth estimation using neural regression forest.
\newblock In {\em CVPR}, 2016.

\bibitem{saxena20083}
A.~Saxena, S.~H. Chung, and A.~Y. Ng.
\newblock 3-d depth reconstruction from a single still image.
\newblock {\em International journal of computer vision}, 76(1):53--69, 2008.

\bibitem{silberman2012indoor}
N.~Silberman, D.~Hoiem, P.~Kohli, and R.~Fergus.
\newblock Indoor segmentation and support inference from rgbd images.
\newblock In {\em ECCV}. Springer, 2012.

\bibitem{wang2015towards}
P.~Wang, X.~Shen, Z.~Lin, S.~Cohen, B.~Price, and A.~Yuille.
\newblock Towards unified depth and semantic prediction from a single image.
\newblock In {\em CVPR}. IEEE, 2015.

\bibitem{wang2016surge}
P.~Wang, X.~Shen, B.~Russell, S.~Cohen, B.~Price, and A.~L. Yuille.
\newblock Surge: Surface regularized geometry estimation from a single image.
\newblock In {\em Advances in Neural Information Processing Systems}, pages
  172--180, 2016.

\bibitem{wang2015designing}
X.~Wang, D.~Fouhey, and A.~Gupta.
\newblock Designing deep networks for surface normal estimation.
\newblock In {\em Proceedings of the IEEE Conference on Computer Vision and
  Pattern Recognition}, pages 539--547, 2015.

\bibitem{xie2016deep3d}
J.~Xie, R.~Girshick, and A.~Farhadi.
\newblock Deep3d: Fully automatic 2d-to-3d video conversion with deep
  convolutional neural networks.
\newblock In {\em European Conference on Computer Vision}, pages 842--857.
  Springer, 2016.

\bibitem{zoran2015learning}
D.~Zoran, P.~Isola, D.~Krishnan, and W.~T. Freeman.
\newblock Learning ordinal relationships for mid-level vision.
\newblock In {\em ICCV}, 2015.

\end{thebibliography}
}

\clearpage

%\textbf{{\LARGE Appendix}}
\includepdf[pages=1-]{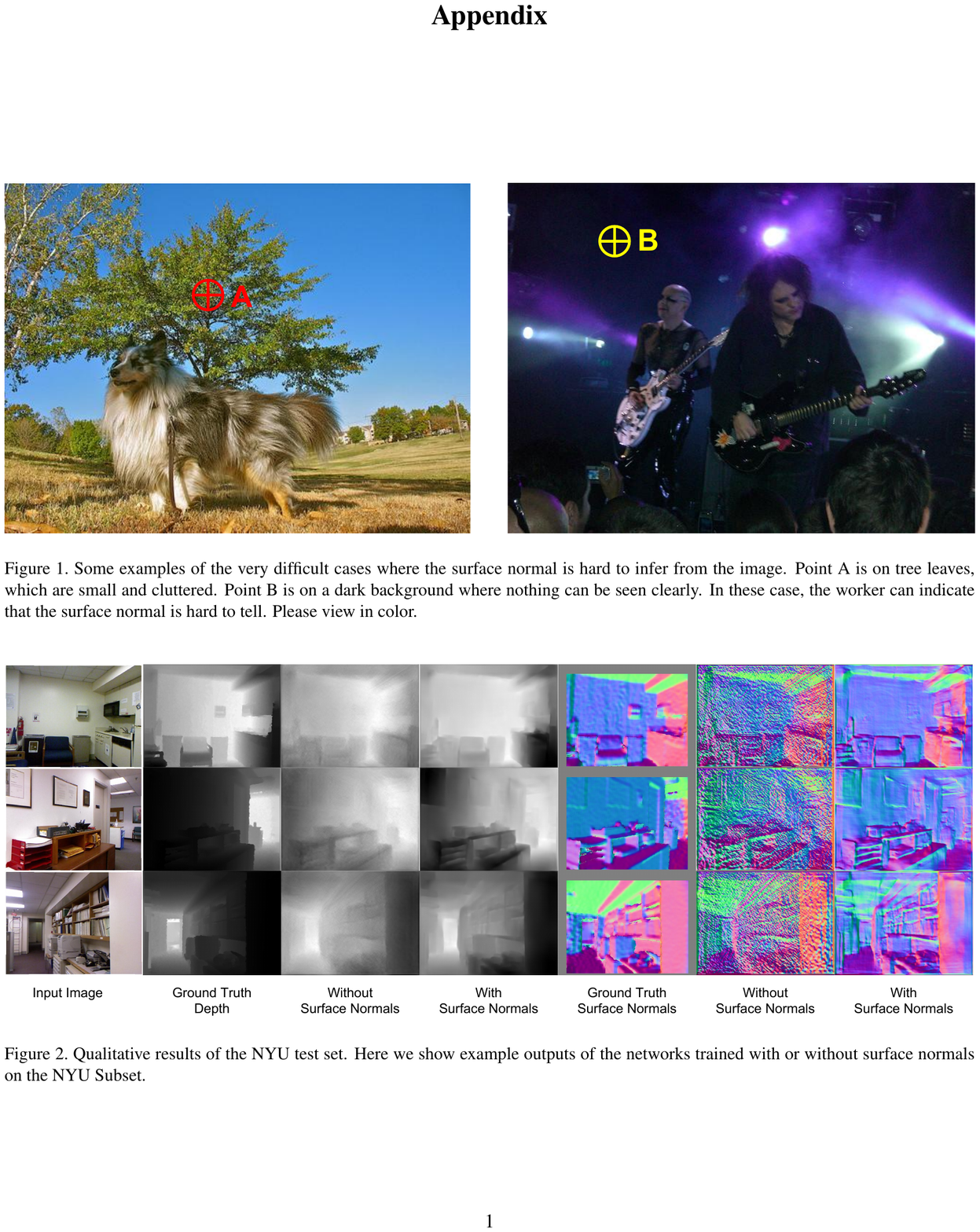}
\includepdf[pages=1-]{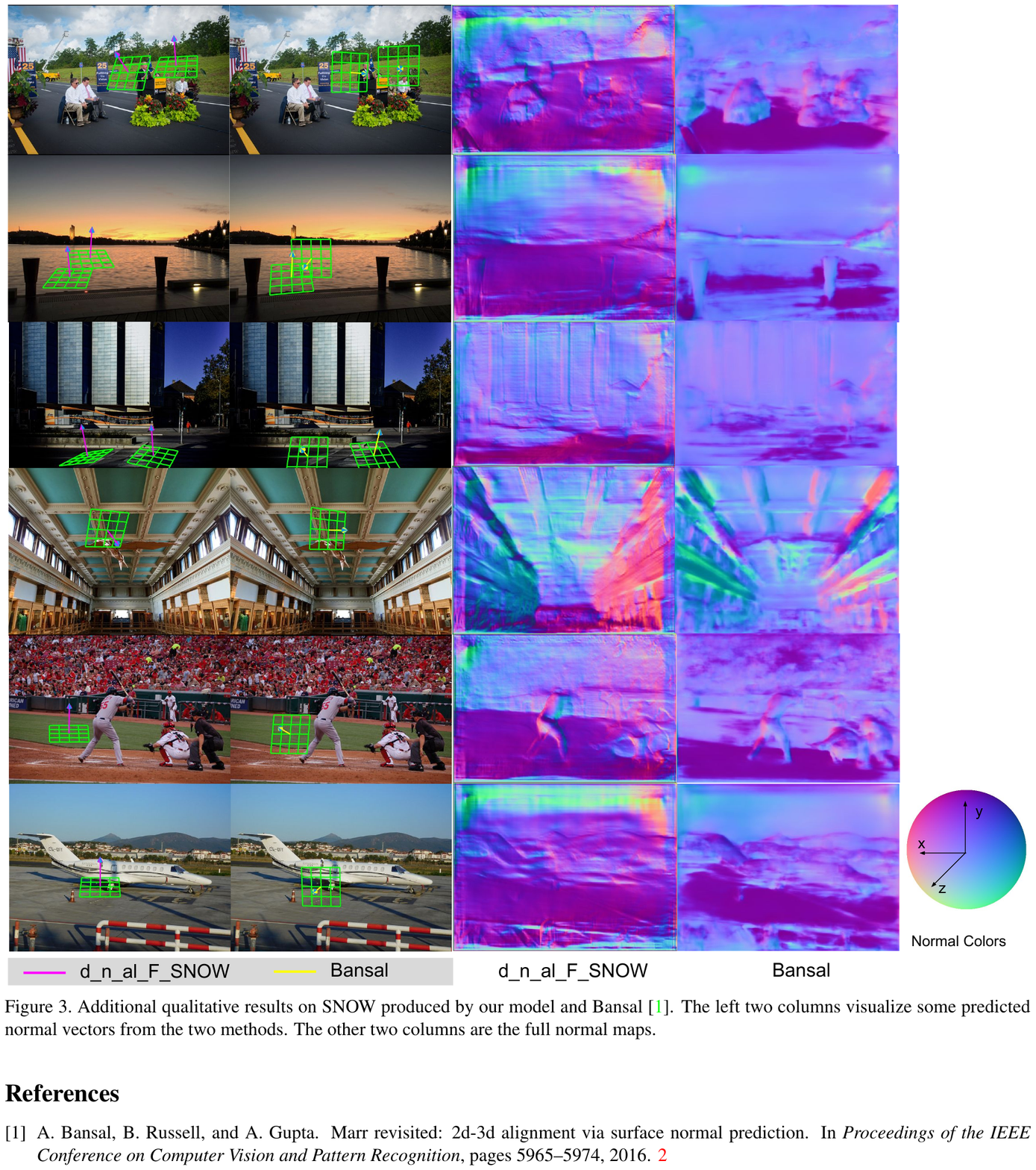}

\end{document}